\documentclass[5p,times]{elsarticle}

\usepackage{amssymb}
\usepackage{algorithm}
\usepackage{algorithmic}
\usepackage[figuresright]{rotating}
\usepackage{subfigure}

\usepackage{amsmath,amssymb}
\usepackage{multirow}

\journal{Expert Systems with Applications}
\begin{document}

\begin{frontmatter}






\title{Sequential Targeting: an incremental learning approach for data imbalance in text classification}


\author[1]{Joel Jang\fnref{label1}}
\ead{wkddydpf@korea.ac.kr}

\author[2]{Yoonjeon Kim}
\ead{yoonkim313@yonsei.ac.kr}

\author[3]{Kyoungho Choi}
\ead{k.h.choi@navercorp.com}

\author[4,5]{Sungho Suh\corref{cor1}}
\ead{s.suh@kist-europe.de}

\address[1]{Department of Computer Science and Engineering, Korea University, 02841 Seoul, Republic of Korea.}
\address[2]{Department of Applied Statistics, Yonsei University, 03722 Seoul, Republic of Korea.}
\address[3]{Media Tech Group, Naver Corp., 13561 Seongnam-si, Gyeonggi-do, Republic of Korea.}
\address[4]{Department of Computer Science, TU Kaiserslautern, 67663 Kaiserslautern, Germany}
\address[5]{Smart Convergence Group, Korea Institute of Science and Technology Europe Forschungsgesellschaft mbH, 66123 Saarbrücken, Germany.} 

\fntext[label1]{This work was done while the author was at NAVER.}
\cortext[cor1]{Corresponding Author.}

\begin{abstract}
Classification tasks require a balanced distribution of data to ensure the learner to be trained to generalize over all classes. In real-world datasets, however, the number of instances vary substantially among classes. This typically leads to a learner that promotes bias towards the majority group due to its dominating property. Therefore, methods to handle imbalanced datasets are crucial for alleviating distributional skews and fully utilizing the under-represented data, especially in text classification. While addressing the imbalance in text data, most methods utilize sampling methods on the numerical representation of the data, which limits its efficiency on how effective the representation is. We propose a novel training method, Sequential Targeting(ST), independent of the effectiveness of the representation method, which enforces an incremental learning setting by splitting the data into mutually exclusive subsets and training the learner adaptively. To address problems that arise within incremental learning, we apply elastic weight consolidation. We demonstrate the effectiveness of our method through experiments on simulated benchmark datasets (IMDB) and data collected from NAVER.
\end{abstract}


\end{frontmatter}


\section{Introduction}
Classification is an important task of knowledge discovery in databases and data mining. It is a task of learning a discriminative function from the given data that classifies previously unseen data to the correct classes. Current research trends in natural language processing focus on developing deep neural network(DNN) models such as BERT \citep{devlin2018bert} that have been pre-trained with a large text corpus and thus show immense improvement in different text classification tasks. Despite the success of large pre-trained models, DNNs still suffer from generalizing to a balanced testing criterion in cases of data imbalance \citep{dong2018imbalanced}. In realistic settings, it is rarely the case where the discrete distribution of the data acquired is perfectly balanced across all classes. Realistic settings are prone to be skewed to specific classes while such classes are often the class of interest. Some situations may be binary, as in detecting spams in forums \citep{ratadiya2019spam}. The majority of the contents posted from users are not spams and is in accordance with the intended goal. As a result, the number of spam samples is sparse in comparison to non-spam samples. Imbalanced data may also occur in a multi-classification setting such as classifying articles into different categories \citep{suh2017comparison}. 

\begin{figure*}
\centering
  \includegraphics[width=0.95\textwidth]{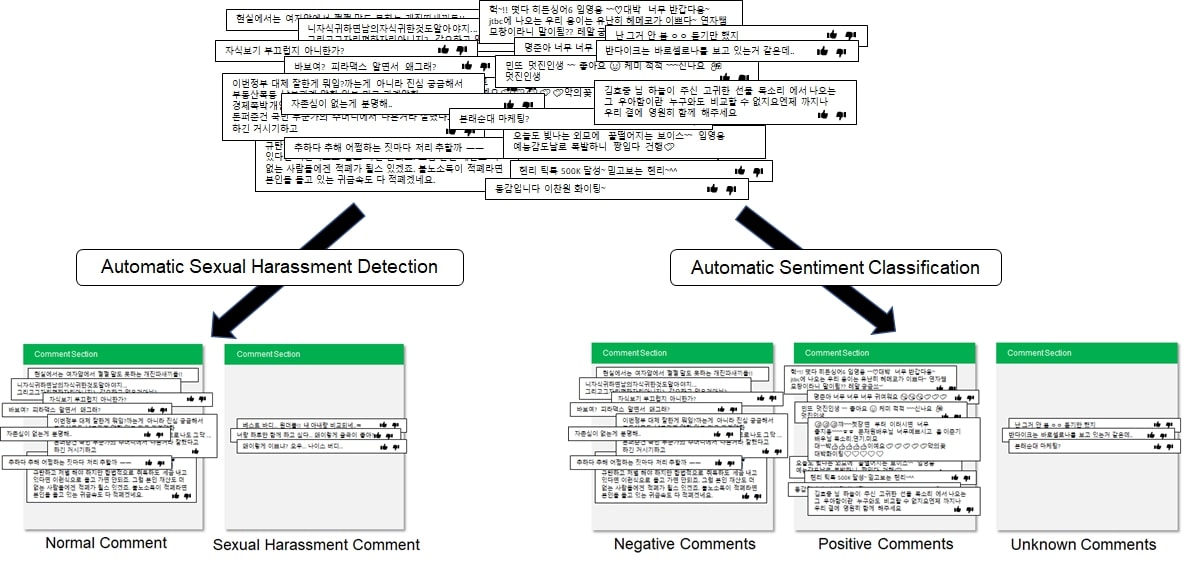}\\
  \caption{Detecting sexual harassment and sentiment in Korean comment data.}\label{fig:comment}
\end{figure*}

Text classification can be used for numerous application purposes. In this paper, we address the problem of detecting sexual harassment and toxicity in comments from news articles. In the name of anonymity, online discussion platforms have become a place where people undermine, harass, humiliate, threaten, and bully others \citep{samghabadi2017detecting} based on their superficial characteristics such as gender, sexual orientation, and age \citep{elsherief2018hate}. Each toxic comment can further be classified into classes based on their degree of toxicity \citep{fortuna2018survey}. Figure \ref{fig:comment} shows the overall procedure of detecting sexual harassment and performing sentimental analysis on comment data in the wild. When collecting and annotating comments, data skewness occurs naturally since users do not consider data imbalance levels when writing toxic or non-toxic comments. Classifiers trained in imbalanced settings tend to become biased toward the class with more samples in the training data. This is because standard deep learning architectures \citep{moon2020beep} do not take the data imbalance level into consideration. In order to develop intelligent classifiers, methods to temper the classifier from biasing towards certain classes are of great importance. 

Previous methods addressing data imbalance in the text can be divided into data-level and algorithm-level methods. Data-level methods \citep{satriaji2018effect, padurariu2019dealing} apply manipulation on the data by undersampling majority classes or oversampling minority classes. However, most of the methods require an effective numerical representation algorithm since methods are applied directly to the representation instead of on the actual text. Algorithm-level methods modify the underlying learner or its output to reduce bias towards the majority group. However, these methods are task-sensitive and somewhat heuristic since it requires the researchers to modify the classifier considering the innate properties of the task. This property leads to the inefficiency in training the learner since heuristic approaches are often time-consuming and arbitrary. Since only traditional oversampling and undersampling methods, which simply duplicate or sample data instances, are independent of these two limitations, methods addressing data imbalance in the text without the utilization of feature spaces or task-dependent is needed.

We propose a novel training architecture, \textbf{Sequential Targeting (ST)}, that handles the data imbalance problem by forcing an incremental learning setting. ST divides the entire training data set into mutually exclusive partitions, target-adaptively balancing the data distribution. Target distribution is a predetermined distributional setting that enables the learner to exert maximum performance when trained with. In an imbalanced distributional setting, the target distribution is idealistic to follow a uniform distribution where all the classes hold equal importance. Optimal class distribution may differ by innate property of the data but research shows that a balanced class distribution has an overall better performance compared to other distributions~\citep{rokach2010ensemble}. The remaining partitions are then sorted in the magnitude of similarity with the target distribution which is measured by KL-divergence. The first partition of the split data is imbalanced while the last partition is arbitrarily modeled to be uniform across classes and all the partitions are utilized to train the learner sequentially. 

We handle the issue of catastrophic forgetting~\citep{french1999catastrophic}, which is an inevitable phenomenon when transfer learning, by utilizing Elastic Weight Consolidation(EWC)~\citep{kirkpatrick2017overcoming} to stabilize the knowledge attained from the previous tasks. This allows the discriminative model to learn from the incoming data while not forgetting the previously inferred parameters from previous tasks. 

Our proposed method is both independent of the numerical representation method and the task at hand. We validate our method on simulated datasets(IMDB) with varying imbalance levels and apply our method to a real-world application. We annotated and construct three datasets consisting of comments made by users from different social platforms of NAVER\footnote{NAVER is the Korean No.1 web search portal where around 16 million users visit every day. \textit{www.naver.com}}: two for detecting sexual harassment and one for multiple sentimental analysis. Annotations on the datasets were improved iteratively by in-lab annotations and crowdsourcing. Experimental results show that ST outperforms traditional approaches, with a notable gap, especially in extremely imbalanced cases. Lastly, ST proves to be compatible with previous approaches.

Our contribution in this paper is three-folds:
\begin{itemize}
  \item We introduce a novel method addressing the data imbalance problem by splitting the data and incrementally training a learner to perform balanced learning by paying equal attention to the minority data as to the majority data. The learner's initial focus on the majority class is compensated by the redistributed task that comes sequentially, resulting in an overall increase of performance. We perform experiments not only on simulated benchmark text datasets but also on datasets constructed from real-world data. Results demonstrate the superiority of our method.
  \item Incremental learning techniques are utilized to prevent catastrophic forgetting on tasks encountered before. The novelty of our method stands since we applied incremental learning to address the data imbalance problem. 
  \item We propose a novel method that is not only free from the dependencies of previous methods addressing the data imbalance in text, but also compatible with the previous methods. The performance increase is shown when utilized together.
\end{itemize}

The rest of the paper is organized as follows. Section \ref{relatedworks} summarizes related works. Section \ref{proposedmethod} provides the details of the proposed method. Section  \ref{experiments}  presents dataset descriptions, experiment setups, and qualitative experimental results on various datasets. Finally, Section \ref{conclusion} concludes the paper.

\section{Related Works}
\label{relatedworks}
\subsection{Methods Handling the Data Imbalance Problem in Text Data}
Previous researchers have proposed data-level methods and algorithm-level methods to address the data imbalance problem. Data-level techniques such as Synthetic Minority Oversampling Technique (SMOTE) \citep{chawla2002smote,han2005borderline} is one of the most used approaches. Its basic idea is to interpolate between the observations of minority classes to oversample the training data. Different variants of SMOTE such as SMOTE-SVM \citep{nguyen2011borderline}and MWMOTE \citep{barua2012mwmote} has also been proposed. Other methods try to augment less-frequently observed classes by translating samples from more-frequently observed classes \citep{kim2020m2m}. Recently, deep learning methods have been applied in imbalanced data setting to generate diverse instances of minority classes by utilizing Generative Adversarial Nets~\citep{goodfellow2014generative, suh2019generative, suh2020cegan}. 

These approaches are non-heuristic but arbitrary to some extent in that researchers should determine what to resample from the data. A classifier should effectively approximate the hyper-plane which forms a boundary between classes. If the resampled data does not fully represent the quality data points that are crucial in deciding the hyper-plane, the model fails to generalize classification performance on new observations. Moreover, most approaches focus on handling imbalance on images, and only a few \citep{satriaji2018effect, padurariu2019dealing} have tried applying SMOTE and its variants to text data. However, these methods assume effective numerical representations of data since SMOTE works only in feature spaces. Even though effective numerical representation methods have been recently proposed such as GLOVE \citep{pennington2014glove} and Word2Vec \citep{bojanowski2017enriching}, methods that can be utilized without having a dependency on the representation method is still in need.

Some text data augmentation methods do not have this limitation and apply augmentation directly on the text data instead of on the numerical representations. Easy data augmentation techniques (EDA) \citep{wei2019eda} introduces four powerful operations: synonym replacement, random insertion, random swap, and random deletion.

Alternately, algorithm-level methods \citep{krawczyk2016learning, khan2017cost}, commonly implemented with a weight or cost schema, modify the underlying learner or its output to reduce bias towards the majority group. Algorithm-level methods modify the structure of the decision process of how much to focus on under-represented samples. This could be implemented by assigning the cost matrix as a penalty~\citep{fernando2017pathnet}. Moreover, loss functions could be modified, such as in the case of focal loss~\citep{lin2017focal}, which reshapes the standard cross-entropy loss such that it penalizes the loss assigned to well-classified instances.

\subsection{Elastic Weight Consolidation} 
Catastrophic forgetting~\citep{french1999catastrophic} is a phenomenon of a learner forgetting about the previously learned task when encountered with another task. This phenomenon occurs naturally when training a learner in an incremental learning setting \citep{li2017learning}. One of the major approaches to overcome this issue is to use an ensemble of networks, each trained with individual tasks \citep{buda2018systematic}. However, this approach has a complexity issue. Alternately, Fernando et al. \citep{fernando2017pathnet} proposed an ensemble approach which attempts to fix the parameters learned from the previous task and train new parameters on consecutive tasks. This method has successfully reduced complexity issues, but performance suffers from the lack of trainable parameters.

Kirkpatrick et al. \citep{kirkpatrick2017overcoming} proposed Elastic Weight Consolidation(EWC) that consider training the neural network from a probabilistic perspective. This approach assumes that neural network parameters follow Gaussian distribution for each task and attempt to find the optimal instance among the distribution that performs well for all tasks. It focuses on developing a regularization term that buffers the model from forgetting the previously trained information. It implements a modified regularization term that consolidates knowledge across tasks by imposing restrictions on the model during training to slow down updating certain important parameters from the previous task. 

EWC is implemented extensively in different applications such as in multi-task learning \citep{rebuffi2018efficient} and mining tasks \citep{karisani2020domain}. Unlike the previous application of EWC which focuses on working in different domains, we apply EWC to address the data imbalance problem.

\section{Sequential Targeting}
\label{proposedmethod}

We first introduce a broad overview of the novel training architecture: Sequential Targeting. Next, we show how this method has been applied to address the data imbalance problem.

\subsection{Matching the Target Distribution with Elastic Weight Consolidation}
\begin{figure}
\centering
  \includegraphics[width=0.95\columnwidth]{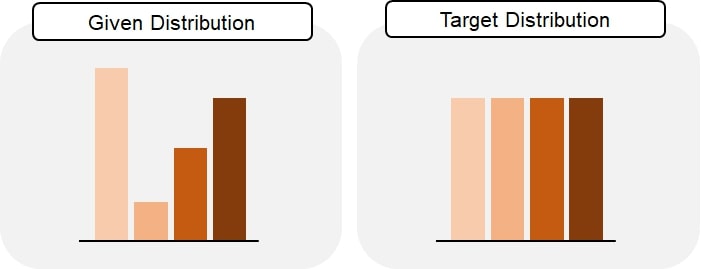}\\
  \caption{An example of when given data distribution differs from the target data distribution}\label{fig:imbalance}
\end{figure}

We propose a novel model architecture of \textbf{forced incremental learning} on the imbalanced setting. The term \textit{forced incremental learning} has rarely, if ever, been used since incremental learning is a much more complicated task. In incremental learning, the new data is referred to as different \textit{task} in this paper. The task is consistently provided therefore the learner needs to be constantly updated on the new task. Because of catastrophic forgetting, learners perform much better when trained with an individual task than continually being updated with multiple tasks. However, by applying EWC, we have proven it is beneficial to force an incremental learning setting where the given data distribution varies substantially from the target data distribution, such as in the case of data imbalance. The \textbf{target distribution}, denoted as $\mathcal{P}_T$, is the idealistic data distribution in which the learner would perform the best if trained on Figure \ref{fig:imbalance} shows an example of a case where the given distribution differs from the target distribution. Left shows the distribution in case of data skewness and right shows the distribution in the data is balanced.

\begin{algorithm}[h]
	\caption{Training architecture for Sequential Targeting}\label{Algo}
	\begin{algorithmic}[1]
		\REQUIRE Given task $\mathcal{D}_{total}$ divided  into multiple tasks, $\mathcal{D}_1, \mathcal{D}_2, ... \mathcal{D}_k$, so that $\bigcup \mathcal{D}_i = \mathcal{D}_{total}$  and $\bigcap \mathcal{D}_i = \emptyset$.
		\STATE \textbf{Initialize:} Sort the tasks $\mathcal{D}_1, \mathcal{D}_2, ... \mathcal{D}_k$ in the decreasing order of Kullback–Leibler divergence from the target distribution $P_{T}$
		\FOR{$i \gets 1$ to $k$}
		\IF{$i=1$}
		\STATE Randomly initialize $\theta_{i}$
		\ELSE
		\STATE Set loss function to $\mathcal{L}(\theta_{i}) + \sum_{j} \frac{\lambda}{2}F_j(\theta_{i,j}  - \theta^{\ast}_{i-1,j})^2$
		\ENDIF
		\STATE Train $\theta_{i}$ from $\mathcal{D}_i$
		\STATE Save optimal $\theta_{i}$ as $\theta^*_i$
		\ENDFOR
	\end{algorithmic}
\end{algorithm}

Our method effectively improves the model by dividing a given task into multiple tasks so that the union of the tasks is the initially given task and the intersection of the divided tasks is an empty set. Each task is partitioned into varying distributions: $\mathcal{P}_1, \mathcal{P}_2, ... \mathcal{P}_k$ where \textit{k} is the number of splits. The learner is sequentially trained on these tasks in the order of similarity with the target distribution, which is measured with KL-divergence. The following has to hold:

\begin{equation}
    \begin{split}
        KL( P_{T} &\Vert P_{prev} )  > KL( P_{T} \Vert P_{curr}),\\
        &\text{where}~P_{i} \xrightarrow{\mathcal D} P_{T}~\text{as}~i~\xrightarrow{}~k.
    \end{split}
    \label{eqn:kl}
\end{equation}

KL divergence is used to measure the discrepancy between the task distributions and the target distribution. 

Using a single learner over all tasks, we incrementally condition the maximum performance from the previous task and stabilize the learned parameters on the current task. Optimizing the network parameters $\theta$ is equivalent to finding their most feasible values given some data \textit{D}. We can compute this conditional probability $p(\theta|D)$ from the prior probability of the parameters $p(\theta)$ and the probability of the data $p(D|\theta)$ by using Bayes’ rule:

\begin{equation}
    \log p(\theta|D) = \log p(D|\theta) + \log p(\theta) - \log p(D).
    \label{eqn:equ1}
\end{equation}

By assuming that the data is split into two independent parts, one defining the previous task $D_{prev}$ and the other current task $D_{curr}$. we can rearrange \eqref{eqn:equ1} by applying the Bayesian update rule:

\begin{equation}
    \resizebox{.91\hsize}{!}{$\log p(\theta|D) = \log p(D_{curr}|\theta) + \log p(\theta|D_{prev}) - \log p(D_{curr})$.}\label{eqn:equ2}
\end{equation}

Equation \eqref{eqn:equ2} shows how the prior distribution learned from the data of the previous task is further enriched by the data given at the current task. 

We apply EWC between transfer learning from task to task. Due to the intractability of the posterior distribution, EWC poses an assumption on the prior distribution to follow a Gaussian distribution with the mean as $\theta^*_{prev}$ and the Fisher information matrix, \textit{F}, of the previous task as the precision. To minimize the loss of information, the objective function is defined as follows:

\begin{equation}
    \mathcal{L}(\theta) = \mathcal{L}_{curr}(\theta_{curr}) + \sum_{i} \frac{\lambda}{2}F_i(\theta_{curr,i}  - \theta^\ast_{prev,i})^2,
    \label{eqn:ewc}
\end{equation}

where $\mathcal{L}_{curr}(\theta)$ sets the loss for the current task, $\lambda$ sets the importance of the previous task and \textit{i} labels each parameter. 

\begin{figure*}
\centering
  \includegraphics[width=0.8\textwidth]{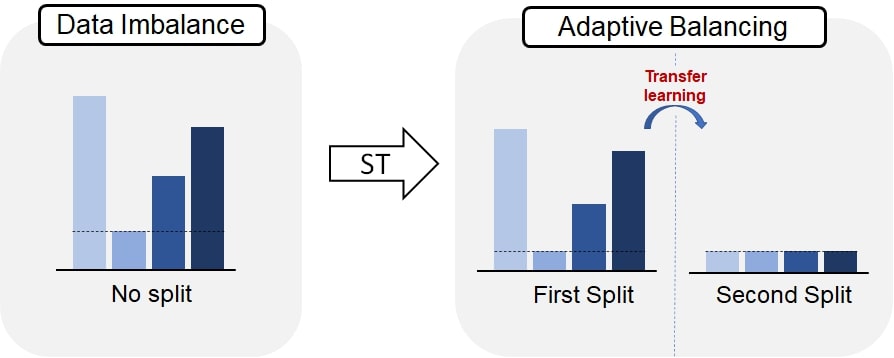}\\
  \caption{Data distribution before and after applying ST in the case of multi-classification with data imbalance.} \label{fig:ST}
\end{figure*}

Since the tasks are sorted dependent on the KL divergence value with the target distribution, the learner is adaptively trained on data distribution similar to the target distribution more and more as tasks proceed. At last, the learner is trained on task that is identical to the target distribution ensuring $\displaystyle KL(P_{T} \Vert P_{k}) \approx 0$. As tasks develop, we apply EWC so that the learner does not forget about previous tasks. The training procedure for the proposed method is summarized in Algorithm \ref{Algo}. 

The number of data splits, \textit{k}, and how each task is partitioned to have varying KL-divergence values are both highly dependent on what $\mathcal{P}_T$ is defined to be. How ST has been applied to address the data imbalance problem is explained in the next subsection.

\subsection{Adaptive Balancing}
Imbalanced data setting tempers the model from learning extensively from under-represented instances. Therefore, it is crucial to enforce the learner to acquire knowledge equally among classes. In an idealistic setting, a learner is trained with equally represented data. However, in a realistic setting, imbalanced data includes some under-represented classes and the learner has difficulty acquiring sufficient knowledge to generalize. ST enables balanced learning by redistributing the task to approach the target distribution as tasks develop. 

It is idealistic to assume $\mathcal{P}_T \,{\buildrel d \over =}\, Uniform[0, p]$ in an imbalanced data setting therefore when \textit{p} denotes the number of classes the target distribution is discrete uniform: $\{\frac{1}{p}, \frac{1}{p}, ,,,, \frac{1}{p}\}$. In our method, the training data is redistributed into two splits so that the last split is identical to the uniform distribution as shown in Figure \ref{fig:ST}. The left shows the data distribution setting before ST is applied and the right shows the case after ST is applied. $\rho$ and $\eta$ are ratios that are explained in Section \ref{experiments}.

This training architecture lets the learner pay more focus to the under-represented data by manipulating the learning sequence. Sequentially training the learner to be exposed to an increasing portion of minority class data benefits the overall performance. Moreover, applying dropout to the layers and implementing EWC during the transfer between tasks proves to help the learner maintain the knowledge acquired from the previous split. We believe this training approach is the first of its kind, with the best of our knowledge. 

\section{Experiments}
\label{experiments}
\subsection{Evaluation Metrics and ratios}
\textit{Accuracy} is commonly used to measure the performance of a classification model. However, when it comes to skewed data, accuracy alone can be misleading and thus other appropriate metrics are needed to correctly evaluate the performance of the model. In this paper, we use precision, recall, and macro F1-score to objectively evaluate the model in a skewed data setting. \textit{Precision} measures the percentage of actual positive among the number of positively predicted samples. \textit{Recall} measures the percentage of the truly positive instances that was correctly predicted by the model. As precision and recall are in a trade-off relationship, selecting a learner that performs well on both metrics would be a reasonable policy. \textit{Macro F1-score} combines both precision and recall as a harmonic mean weighted with equal importance on each class rather it be sparse or rich. In this paper, F1-score is used as the core metric for measuring performance.

Following the conventions of the previous research on imbalanced data \citep{buda2018systematic,johnson2019survey}, we employ two distinct ratios used throughout the experiment to represent the imbalanced state of the data. One parameter is a ratio between the number of instances in majority classes and the number of instances in minority classes and defined as follow:

\begin{equation}
   \rho = \frac{|\max(C_i)|}{|\min(C_i)|}.
   \label{eqn:rho}
\end{equation}

where $C_{i}$ is the set of instances in class \textit{i} and \textit{N} is the total number of classes. Another 

The other parameter $\eta$ is a parameter that compares the relative number of minority class instances among splits. For instance, if the first task consists of 100 samples in minority class and the second task consists of 50 samples, then $\eta$ will be 2:1. We experienced various combinations of these three ratios and concluded that $\eta$ does not have a significant effect on model performance. Therefore, the number of instances of the minority classes between the splits is fixed as 1:1 throughout the whole experiment.

\subsection{Dataset Descriptions}
In this paper, we validate and apply our proposed method on four text datasets(IMDB, NAVER-Catcall-Yellow, NAVER-Catcall-Red, NAVER-Posneg). In the case of \textit{IMDB}, the dataset was deliberately made into varying imbalanced states as shown in Table \ref{tab:textdataset}. \textit{NAVER-Posneg, NAVER-Catcall-Yellow, and NAVER-Catcall-Orange} are Korean text data of comments crawled and annotated to improve the AI Clean Bot 2.0~\citep{Lee2020cleanbot} used to detect toxic comments at NAVER. All of the data were collected from the news, entertainment, and sports platforms of NAVER. Annotations on the datasets were made through crowdsourcing and improved iteratively by in-lab annotations. All of the datasets have been divided into train, validation, and test sets for objective comparison of methods.

\begin{table*}[ht!]
    \centering
    \caption{Simulated IMDB Datasets with Variations of $\rho$ ratio and NAVER Datasets.}
    \begin{tabular}{c|c|cc|cc|cc}
        \hline\hline
        \multirow{2}{*}{Dataset} & \multirow{2}{*}{$\rho$} & \multicolumn{2}{c}{Train} & \multicolumn{2}{c}{Validation} & \multicolumn{2}{c}{Test} \\ \cline{3-8} 
        & & Minority    & Majority    & Minority       & Majority      & Minority    & Majority   \\ \hline
        \multirow{3}{*}{IMDB} & 10 &1,250 & 12,500 &2,500 & 2,500 &10,000 &10,000 \\ \cline{2-8} 
        & 20 & 625 & 12,500 & 2,500 & 2,500 & 10,000 & 10,000\\ \cline{2-8} 
        & 50 &250 & 12,500 & 2,500 & 2,500 & 10,000 & 10,000 \\ \hline
        NAVER-Posneg & 8 & 6,355 & 48,342 & 314 & 2,498 & 621 & 1453\\ \hline
        NAVER-Catcall-Yellow & 10 & 2,650 & 26,792 & 140 & 1,410 & 73 & 427\\ \hline
        NAVER-Catcall-Red & 15 & 1,840 & 27,602 & 97 & 1,453 & 73 & 427\\ \hline\hline
    \end{tabular}
    \label{tab:textdataset}
\end{table*}

\textbf{IMDB}~\citep{maas2011learning} is a text dataset, which contains 50,000 movie reviews with binary labels (positive/negative). Reviews have been preprocessed, and each review is encoded as a sequence of word indexes. Three different imbalance ratios have been deliberately made ($\rho=10,20,50$) to test how each method performs as the imbalance level worsens. The positive reviews are regarded as the positive class.

\begin{figure}
    \centering
    \includegraphics[width=.99\columnwidth]{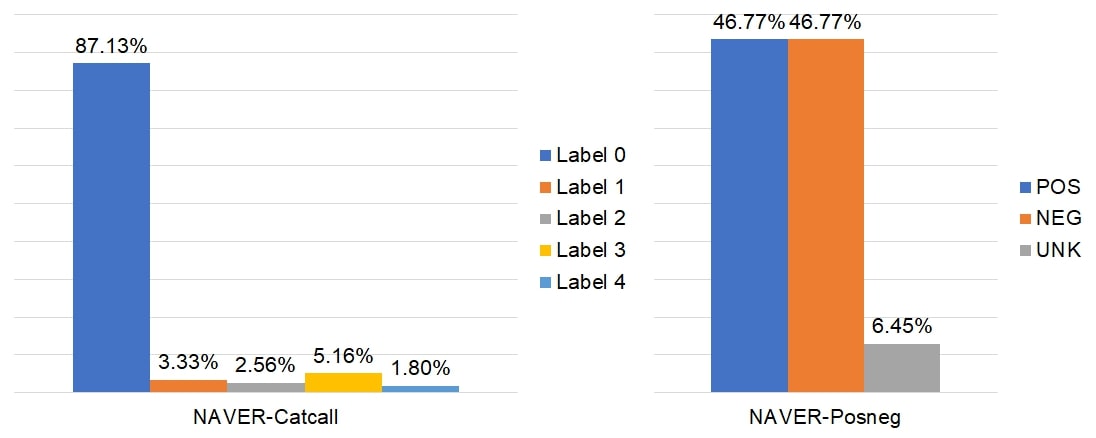}\\
    \caption{Initial data distribution of collected and annotated comments of \textit{NAVER-Catcall} and \textit{NAVER-Posneg}.}\label{fig:naverdist}
\end{figure}

\textbf{NAVER-Catcall-Yellow and NAVER-Catcall-Red} consists of comments that have been collected and annotated to train a model that detects sexual harassment in comments. It is made up of 31,392 comments labeled into 5 different classes. The labeling rules used for the annotation process and how the initial annotated data was used to construct \textit{NAVER-Catcall-Yellow} and \textit{NAVER-Catcall-Red} are shown in Table \ref{tab:catcallrule}. The second column shows the initial labeling by annotators. The third and fourth column shows the binary labeling standards for \textit{yellow} and \textit{red}. As shown, models trained with \textit{yellow} standard was expected to become more sensitive since the initial \textit{label 2} was considered a positive case of sexual harassment. Likewise, models trained with \textit{red} standard was expected to have lower recall, but show higher precision since it only perceives clear sexual harassment comments as positive. The left bar graph in Figure \ref{fig:naverdist} shows the initial data distribution and the percentage of each class labeled by the annotators.

\begin{table}[ht!]
	\caption{Labeling rules used to construct NAVER-Catcall-Yellow and NAVER-Catcall-Red.}
	\label{tab:catcallrule}
	\centering
	\resizebox{\columnwidth}{!}{%
	\begin{tabular}{|l|c|c|c|}
		\hline
		\textbf{Description} & \textbf{Label} & \textbf{Yellow} & \textbf{Red} \\\hline
		No one will perceive it as sexual harassment & 0 & \multirow{2}{*}{0} & \multirow{3}{*}{0}\\ \cline{1-2} 
		Contains sexual language & 1 & & \\ \cline{1-2} \cline{3-3} 
		Some may perceive as sexual harassment & 2 &\multirow{3}{*}{1} &\\ \cline{1-2} \cline{4-4} 
		Clearly intended as sexual harassment & 3 & &\multirow{2}{*}{1}\\ \cline{1-2} 
	    Shows advocacy of sexual violence & 4 & &\\ \hline
	\end{tabular}
	}
\end{table}

The actual analysis of imbalance found on real comments was much severe ($\rho > 100$) since sexual harassment comments were not common. Because of this, we crawled comments from media sources where sexual harassment comments were more likely to be written. Despite sampling from less imbalanced data sources, the data imbalance level still turned out to be significant: $\rho\approx10$ for \textit{yellow} and $\rho\approx15$ for \textit{red}.

\textbf{NAVER-Posneg} was collected to train a sentimental analysis model and consists of 95,875 comments labeled into three different classes: \textit{positive, negative, and unknown}. The \textit{unknown} class was added since some comments were hard to label either as \textit{positive} or \textit{negative} based on the comment itself without knowing the full context. \textit{Positive} and \textit{negative} comments were proportionate. However, skewnewss existed in \textit{unknown} comments ($\rho\approx8$). The right bar graph in Figure \ref{fig:naverdist} shows the data distribution of the constructed dataset. 

\subsection{Experimental Setup}
In our experiments, our proposed method has been extensively compared with two data-level methods, random oversampling (ROS) and random under-sampling (RUS). For ROS, instances from the minority classes were sampled and duplicated to match the number of instances in the majority classes. For RUS, we randomly down-sampled all majority classes so that they had the same number of instances as the minority classes. 

We explore the full capability of ST by combination with ROS. ROS is combined with ST by oversampling the first split; no sampling method is applied to the second split. In the case of \textit{NAVER-Catcall-Red}, we perform experiments with EDA \citep{wei2019eda} as well as combine EDA with ST by applying EDA to the first split. We only utilize synonym replacement and random insertion for EDA.

Even though large models, such as BERT~\citep{devlin2018bert}, tend to perform better in imbalanced settings, a simple CNN+LSTM model architecture was used since large models suffer from slow inference time \citep{li2020train}. Inference time and model size are important factors when considering actual model deployment that will be used in applications. The details of the model structure used for experiments are shown in Table \ref{tab:textmodel}. While BERT has 110M parameters, our model has only 1M parameters (\textit{more than 100 times smaller}) and thus show faster inference time.

At the last FC layer, sigmoid activation is used for binary classification and softmax activation is used for multi-classification. The CNN layer efficiently extracts the higher-level representations while the BiLSTM and UniLSTM layers obtain the sentence representation \citep{zhou2015c}.

\begin{table}[ht!]
	\caption{Neural Network Architecture of the CNN+LSTM classifier}
	\label{tab:textmodel}
	\centering
	\resizebox{\columnwidth}{!}{%
	\begin{tabular}{lcc}
		\hline
		\textbf{Layers} & \textbf{Activation function} \\\hline\hline
		Embedding & - \\ 
		1-Dimension Convolution Layer (Dropout: 20\%) & ReLU \\
		BiLSTM + Residual Connection from Embedding & -\\ 
		UniLSTM (Dropout: 20\%) & - \\ 
		1-Dimension Global Max Pooling & - \\ 
		1-Dimension Fully Connected & Sigmoid/Softmax \\ 
		\hline
	\end{tabular}
	}
\end{table}

For each configuration, five independent trial runs were made with different initial weights. This setting ensures the effect of weight initialization to be ruled out in evaluating model performance. Among the five trials, the model checkpoint with the highest validation f1-score was used for evaluation. All the experimental settings including epochs, learning rate, and model architecture are fixed for each corresponding task. 

\subsection{Experiment Results}

\begin{table*}[ht!]
    \caption{Experimental Results on Simulated IMDB Datasets with Varying $\rho$ ratio}
    \centering
    \begin{tabular}{c|ccc|ccc|ccc}
    \hline\hline
    Ratios        & \multicolumn{3}{c|}{$\rho=10$} & \multicolumn{3}{c|}{$\rho=20$} & \multicolumn{3}{c}{$\rho=50$}\\ \hline
    Metrics        & \multicolumn{1}{c|}{F1} & \multicolumn{1}{c|}{Precision} & \multicolumn{1}{c|}{Recall} & \multicolumn{1}{c|}{F1} & \multicolumn{1}{c|}{Precision} & \multicolumn{1}{c|}{Recall} & \multicolumn{1}{c|}{F1} & \multicolumn{1}{c|}{Precision} & \multicolumn{1}{c}{Recall} \\ \hline
    Baseline        & 0.7291 & \textbf{0.8751} & 0.6248 & 0.3216 & 0.6677 & 0.2118 & 0.0833 & 0.69 & 0.0443 \\ 
    ROS            & 0.7548 & 0.702 & 0.8162 & 0.5738 & 0.5839 & 0.564 & 0.5837 & 0.5618 & 0.6061\\ 
    RUS            & 0.7921 & 0.7305 & 0.865 & 0.1837 & 0.6758 & 0.1063 & 0.0256 & 0.6065 & 0.0131 \\ 
    ST       & 0.8002 & 0.7181 & \textbf{0.9035} & \textbf{0.7543} & \textbf{0.7368} & 0.7727 & 0.6457 & \textbf{0.7188} & 0.586 \\ 
    ST + ROS & \textbf{0.8143} & 0.7984 & 0.8307 & 0.7259 & 0.6728 & \textbf{0.788} & \textbf{0.6471} & 0.6156 & \textbf{0.6819} \\ \hline\hline
    \end{tabular}
    \label{tab:imdb}
\end{table*}
    
\begin{figure}
    \centering
    \includegraphics[width=.99\columnwidth]{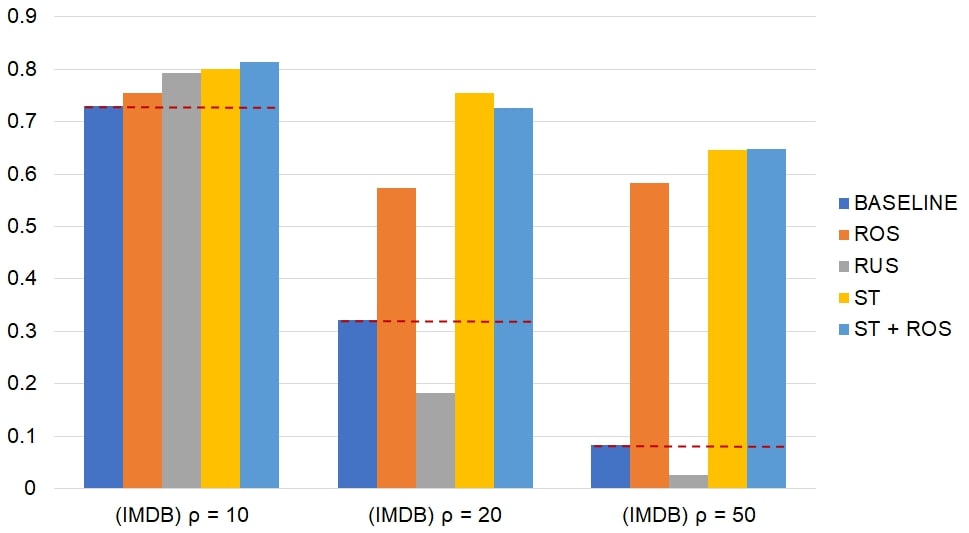}\\
    \caption{Experimental Results on IMDB Datasets in terms of F1-Score.}\label{fig:imdb}
\end{figure}

\textbf{IMDB Results.} Table \ref{tab:imdb} shows the experimental results on the simulated IMDB datasets. Baseline methods show a substantial performance decrease when the data imbalance level worsens. This is because of the significant decrease in the recall, which signifies only a small portion of positive reviews were actually classified correctly. We observe training the model with ST outperforms, if not on par with, traditional methods. Results show a considerable increase in recall when ST is applied. This is because the model can classify more positive reviews correctly since the focus was put on the under-represented class. It is natural to expect a significant decrease in precision since the focus has been shifted away from the majority class. However, the drop is minimal when EWC is applied. This is because EWC helps the model to remember valuable information obtained during the training of the first split as the model is trained with the balanced second split. As shown in Figure \ref{fig:imdb}, the performance gap between ST and baseline grows substantially as the severity of data imbalance increases: 8.52\%, 43.27\%, and 56.38\%.

\begin{table*}[ht!]
    \caption{Experimental Results on NAVER-Catcall-Yellow and NAVER-Posneg}
    \centering
    \begin{tabular}{c|ccc|ccc}
    \hline\hline
    Ratios        & \multicolumn{3}{c|}{(NAVER-Catcall-Yellow)~$\rho=10$} & \multicolumn{3}{c}{(NAVER-Posneg)~$\rho=8$} \\ \hline
    Metrics        & \multicolumn{1}{c|}{F1} & \multicolumn{1}{c|}{Precision} & \multicolumn{1}{c|}{Recall} & \multicolumn{1}{c|}{F1} & \multicolumn{1}{c|}{Precision} & \multicolumn{1}{c}{Recall} \\ \hline
    Baseline        & 0.5455 & \textbf{0.8108} & 0.411 & 0.7249 & \textbf{0.7756} & 0.7288 \\ 
    ROS            & 0.6218 & 0.8043 & 0.5068 & 0.7126 & 0.7754 & 0.7182 \\ 
    RUS            & 0.5733 & 0.5857 & 0.5616 & 0.7363 & 0.7444 & 0.7353 \\ 
    ST        & \textbf{0.7144} & 0.6974 & 0.726 & \textbf{0.7489} & 0.7651 & \textbf{0.7489} \\ 
    ST + ROS & 0.7018 & 0.6122 & \textbf{0.8219} & 0.7352 & 0.7609 & 0.7365\\ \hline\hline
    \end{tabular}
    \label{tab:naver}
\end{table*}

\begin{figure}
    \centering
    \includegraphics[width=8cm]{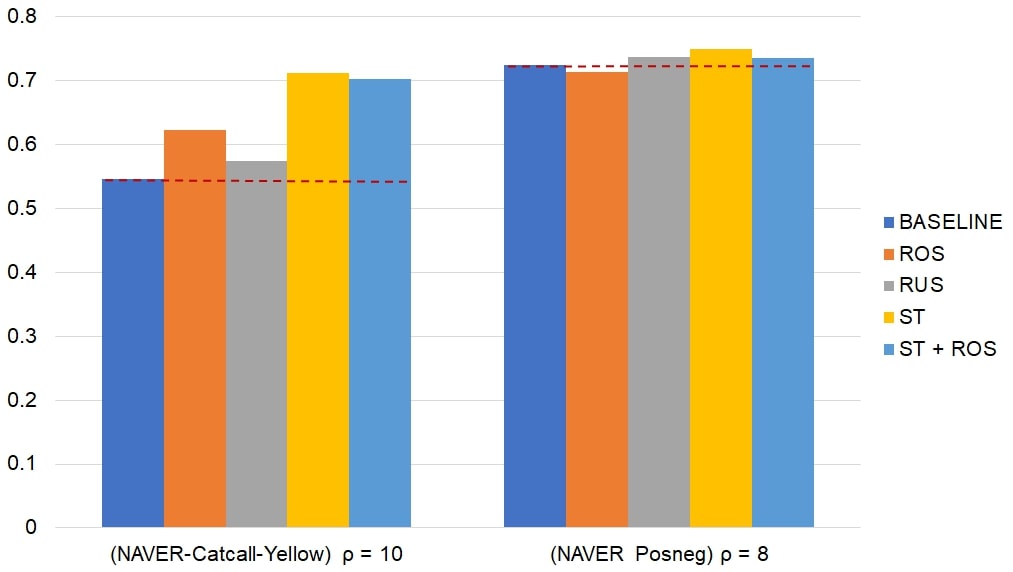}\\
    \caption{Experimental Results on NAVER-Catcall-Yellow and NAVER-Posneg in terms of F1-Score.}\label{fig:naver}
\end{figure}

\textbf{NAVER-Catcall-Yellow and NAVER-Posneg Results.} Table \ref{tab:naver} shows the experimental results on \textit{NAVER-Catcall-Yellow} and \textit{NAVER-Posneg}. In both cases, ST shows the best performance. Applying ST in \textit{NAVER-Catcall} and \textit{NAVER-Posng} shows a performance increase of 16.89\% and 2.4\% in terms of F1-score, respectively. Applying ST has a much more significant impact in the case with only \textit{NAVER-Catcall} than in the case of \textit{NAVER-Posneg}, as shown in Figure \ref{fig:naver}. This is because the former is a binary classification task and the latter is a multi-classification task. Considering that most methods are ineffective and even may cause a negative effect on multi-classification tasks \citep{zhou2005training}, the relatively small increase in performance is still meaningful. 
    
\begin{table}[ht!]
    \caption{Experimental Results on NAVER-Catcall-Red}
    \centering
    \begin{tabular}{c|ccc}
    \hline\hline
    Ratios        & \multicolumn{3}{c}{(NAVER-Catcall-Red)~$\rho=15$}\\ \hline
    Metrics        & \multicolumn{1}{c|}{F1} & \multicolumn{1}{c|}{Precision} & \multicolumn{1}{c}{Recall} \\ \hline
    Baseline        & 0.411 & \textbf{0.9091} & 0.274\\ 
    RUS            & 0.5893 & 0.8462 & 0.4521\\ 
    ROS            & 0.6316 & 0.878 & 0.4931\\ 
    EDA \citep{wei2019eda}  & 0.6917 & 0.7667 & 0.6301\\ 
    ST             & 0.671 & 0.7015 & 0.6438\\ 
    ST + ROS       & 0.7273 & 0.8136 & 0.6575\\
    ST + EDA \citep{wei2019eda} &\textbf{0.7586} & 0.7639 & \textbf{0.7534}\\
  \hline\hline
    \end{tabular}
    \label{tab:naver2}
\end{table}

\begin{figure}
\centering
  \includegraphics[width=7cm]{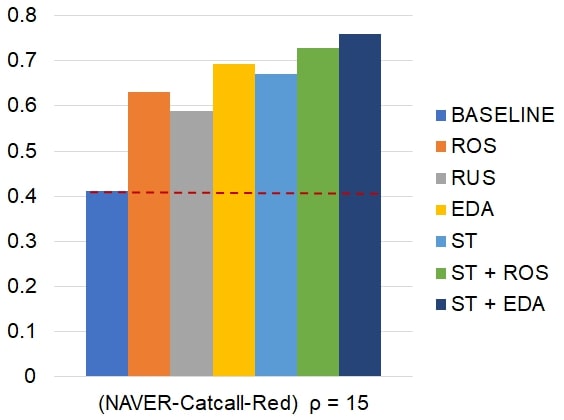}\\
  \caption{Experimental Results on NAVER-Catcall-Red in terms of F1-Score.}\label{fig:naver2}
\end{figure}

\textbf{NAVER-Catcall-Red Results}
Table \ref{tab:naver2} shows the experimental results on \textit{NAVER-Catcall-Red}. EDA shows a higher F1-Score than ROS. This is because instead of a plain duplication approach for oversampling, EDA utilized different operations that allowed data to be oversampled similar to actual data in the wild. The effect is emphasized when used together with ST as shown in Figure \ref{fig:naver2}. ST+EDA shows the best performance among compared methods and shows an increase of 34.76\% in terms of F1-score compared to the Baseline. 

\section{Conclusion}
\label{conclusion}
It is seldom the case data in the wild has a balanced distribution. In realistic settings, there is a limitation of acquiring relatively balanced data through choices of balanced data sources. Handling data skewness is a crucial problem because learning from imbalanced data inevitably brings bias toward frequently observed classes. Data-level manipulation tries to under-sample the majority classes or over-sample the minority classes. But these methods tend to discard valuable information from observations of majority classes or overfit to a sparse representation of minority classes, especially as the imbalance level gets higher. Moreover, recent methods such as SMOTE cannot be applied directly to the text data. 

We propose ST, which effectively circumvents these issues by simply decomposing the data into \textit{k} splits and sequentially training a learner in the decreasing order of KL divergence with the target distribution, which in the case of data imbalance problem is the discrete uniform distribution. Through extensive experiments, we show our architecture proves to be compatible with previous methods and outperforms existing methods when validated on simulated as well as real-application tasks. Our model shows superiority in performance because it enables more focus to be put on minority instances while not forgetting about majority instances. We believe that our work makes a meaningful step towards handling data skewness in text classification and the application of incremental learning methods focused on the data imbalance problem.

For future work, ensemble methods can be used by varying the $\eta$ ratio to train multiple weak learners. Moreover, since ST can be applied simultaneously with algorithm-level methods, proven methods such as focal loss \citep{lin2017focal} and cost-sensitive deep neural network \citep{khan2017cost} could be implemented together to increase optimal performance.

\section*{Acknowledgement}
This research work is supported by NAVER Corp.

\bibliographystyle{model5-names}\biboptions{authoryear}
\bibliography{article}







\end{document}